\documentclass{article}

\usepackage{arxiv}

\usepackage[utf8]{inputenc} 
\usepackage[T1]{fontenc}    
\usepackage{hyperref}       
\usepackage{url}            
\usepackage{booktabs}       
\usepackage{amsfonts}       
\usepackage{nicefrac}       
\usepackage{microtype}      
\usepackage{lipsum}
\usepackage{graphicx}
\usepackage{xcolor}
\usepackage{todonotes}
\usepackage{amsmath}
\usepackage{graphicx,subcaption,lipsum}
\usepackage{wrapfig}
\title{Curriculum Learning for Mesh-based simulations}

\author{
 Paul Garnier and Vincent Lannelongue\thanks{Authors are ordered alphabetically and not by contributions.} \\
 \textbf{Elie Hachem} \\
Mines Paris - PSL University\\
  Centre for Material Forming (CEMEF) \\
  CNRS \\
  \texttt{firstname.surname@minesparis.psl.eu} \\
}

\begin{document}
\maketitle
\begin{abstract}
Graph neural networks (GNNs) have emerged as powerful surrogates for mesh-based computational fluid dynamics (CFD), but training them on high-resolution unstructured meshes with hundreds of thousands of nodes remains prohibitively expensive. We study a \emph{coarse-to-fine curriculum} that accelerates convergence by first training on very coarse meshes and then progressively introducing medium and high resolutions (up to \(3\times10^5\) nodes). Unlike multiscale GNN architectures, the model itself is unchanged; only the fidelity of the training data varies over time. We achieve comparable generalization accuracy while reducing total wall-clock time by up to 50\%. Furthermore, on datasets where our model lacks the capacity to learn the underlying physics, using curriculum learning enables it to break through plateaus.
\end{abstract}

\section{Introduction}
\label{sec:introduction}

High-fidelity CFD solvers based on the Navier--Stokes equations underpin critical engineering pipelines, from aerodynamic shape design to climate modeling. Unfortunately, their computational cost scales steeply with mesh resolution, especially on complex three-dimensional geometries. Graph neural networks (GNNs) have recently been adopted as surrogates that operate directly on unstructured meshes and can predict flow fields orders of magnitude faster than traditional solvers \cite{pfaff2021learning}. However, training these models on the \emph{same} fine meshes used in production remains very challenging due to time and memory limitation.

A natural strategy is to exploit the multiresolution structure of CFD datasets: coarse meshes capture large-scale flow features at a fraction of the cost. In contrast, fine meshes resolve small vortices and boundary layers. Inspired by \emph{curriculum learning} \cite{bengio} and its recent incarnations in language modeling \cite{nagatsuka-etal-2021-pre, pouransari2025datasetdecompositionfasterllm} and computer vision via progressive image resizing \cite{wang2023efficienttrainexploringgeneralizedcurriculum}, we investigate how a \textbf{data-centric curriculum} can ease training for mesh-based GNN surrogates. 

Specifically, we keep the GNN architecture and optimizer intact but vary the fidelity of samples seen during training: the network is first exposed only to very coarse meshes, then to medium resolutions, and finally to the original high-resolution meshes, similarly to \cite{lam2023graphcastlearningskillfulmediumrange,garnier2025trainingtransformersmeshbasedsimulations}.

This is distinct from multigrid or multiscale architectures that simultaneously process multiple graph resolutions \cite{lino2021simulating, Yang2022amgnet, cao2023efficientlearningmeshbasedphysical, taghibakhshi2023mggnn, garnier2025multi}, or approaches that interpolate fine resolution data to coarse ones \cite{fortunato2022multiscalemeshgraphnets}. We aim to go further than previous studies from \cite{song2021transferlearningmultifidelitydata, liu:hal-03878200, Taghizadeh2024} and compare different strategies in terms of computation time, FLOPs, minimal mesh size, and final accuracy.

\section{Datasets}
\label{sec:datasets}

\begin{figure}[ht!]
  \makebox[\textwidth][c]{\includegraphics[width=1\textwidth]{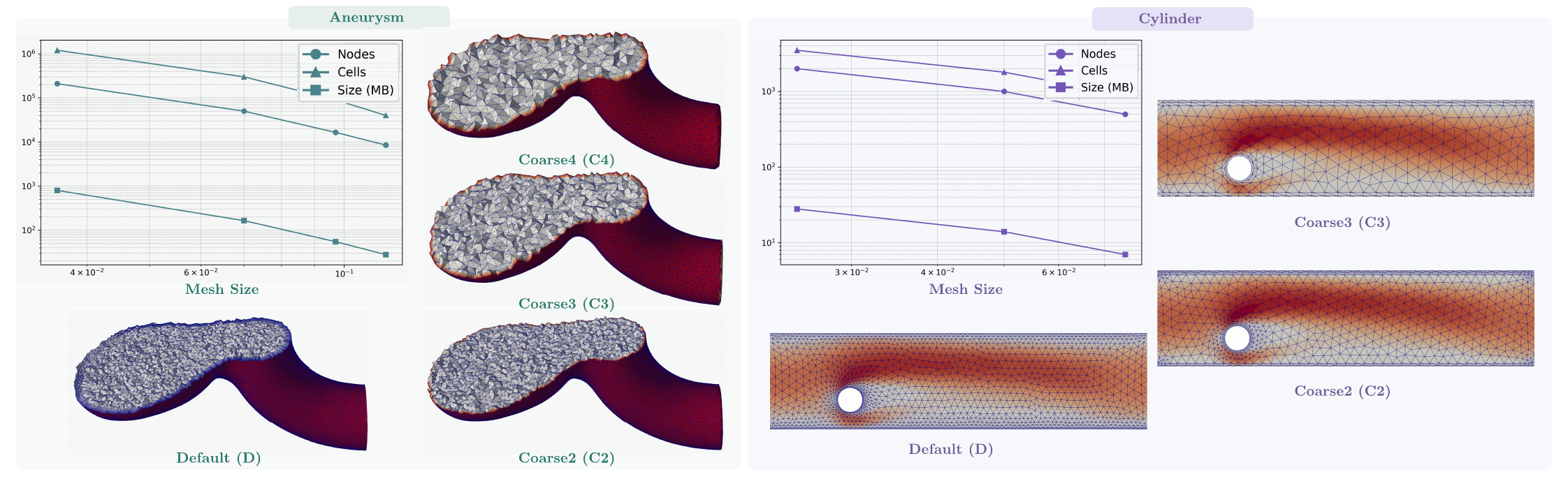}}%
  \caption{Presentation of the two datasets used in our experiments. \textbf{(left)} Mesh size used for each version of the datasets, and the associated number of nodes, cells, and size in MB on average per mesh trajectory. \textbf{(right)} Meshes from the datasets, using a crinkle slice in the middle of the geometry to display the vertices inside the aneurysm. From most coarse to fine.}
  \label{fig:meshsize}
\end{figure}

We perform our experiments on two different datasets. First, \cite{goetz_anxplore_2024} developed 101 semi-idealized geometries derived from patient-specific intracranial aneurysms, segmented from medical imaging data. They conducted CFD simulations of blood flow within these vessels, numerically solving the transient incompressible Navier-Stokes equation over a complete cardiac cycle. This leads to 101 trajectories, each of 79 time steps (with $\Delta t = 0.01s$), where the meshes consist of roughly 300,000 nodes and 3 to 4 million edges. Second, we use the Cylinder benchmark in 2D from \cite{pfaff2021learning}. We use 100 training trajectories of 600 time steps (with $\Delta t = 0.01s$) where meshes consist roughly of 2,000 nodes and 4,000 edges. An overview of the datasets is available in \autoref{fig:meshsize}.

\subsection{Coarsening methodology}
\label{subsec:methodo}

The aneurysm dataset was produced through an automated pipeline. Starting from the high-resolution CFD meshes with refined boundary layers, we first extracted the outer surfaces of each vascular geometry and re-meshed them using isotropic triangular elements. These surface meshes were then used to construct volumetric meshes of unstructured tetrahedral elements with uniform resolution, adopting average target element sizes of 0.3 mm for the first coarsening phase. To effectively reduce the number of elements while keeping a reasonable element size inside the domain, no boundary layer refinement was included in the lighter datasets. We repeat this operation three times in total to obtain three coarser datasets.

Similarly, the Cylinder dataset meshes have refined boundary layers near the top and bottom walls and the cylinder, capturing accurate fluid velocity gradients while maintaining a no-slip wall condition. Since the size of these meshes is significantly smaller, we decided to retain their local refinements. To coarsen these anisotropic meshes, we augmented the minimum and maximum element sizes and increased the thickness of the boundary layers. This allowed us to generate meshes of coarse and medium resolutions while preserving the local refinement at the wall and cylinder boundaries. This operation is applied twice to obtain two coarser datasets.

Flow fields were transferred onto these meshes by linearly interpolating velocity data from the high-resolution CFD simulations for both datasets. 

\section{Training protocol}
\label{sec:training}

\subsection{Model}
\label{subsec:model}

We introduce our transformer-based GNN\footnote{We reproduced our experiments using a Message Passing architecture with 15 layers of width 128 and find the same trends.}, a 500k-parameter model with $L=10$ layers processing tokens in $\mathbb{R}^{d=64}$ from \cite{garnier2025trainingtransformersmeshbasedsimulations}. Our model takes a graph $G_t$ as input and predicts $G_{t+1}$. Graphs only hold per-node features, such as the velocity, the acceleration, the node's position or its type. We ensure the boundary conditions on the vessels and aneurysm walls by setting the velocity field to 0 at those locations, and apply a similar strategy to the cylinder. 

Our model follows an Encode-Process-Decode architecture with a Transformer processor. The architecture works by first encoding the node features in a latent space $\mathbf{Z} = (\mathbf{z}_1, \mathbf{z}_2,...,\mathbf{z}_N)^{T} \in \mathbb{R}^{N \times d} = \mathcal{E}(G)$. We then process $\mathbf{Z}$ through $L$ Transformer Blocks. Each block takes the latest latent representation $\mathbf{Z}$ and the Adjacency Matrix $\mathbf{A}$ as input:

\begin{align}
    \mathbf{Z'}_l &= \text{RMSNorm}\big(\text{MMHA}(\mathbf{Z}_{l-1}, \mathbf{A}) + \mathbf{Z}_{l-1}\big) && \ell \in [1\ldots L] \label{eq:msa_apply} \\
    \mathbf{Z}_l &= \text{RMSNorm}\big(\text{GatedMLP}(\mathbf{Z'}_{l}) + \mathbf{Z'}_{l}\big) && \ell \in [1\ldots L] \label{eq:mlp_apply}
\end{align}

$\text{MMHA}$ is a Multi-Head attention operation masked by the Adjacency matrix:

\begin{equation}
    \text{Attention}(\mathbf{Z}) = \Bigg(\mathbf{A}\odot\sigma\Big( \frac{QK^T}{\sqrt{d}} \Big)\Bigg)V
\end{equation}

where $Q, K, V$ are linear projections of $\mathbf{Z}$.

$\text{GatedMLP}$ is a gated Multi-layer perceptron following the work of \cite{dauphin2017language}:

\begin{equation}
    \mathbf{Z} = W_f\Big(\text{GeLU}\big(W_l \mathbf{Z} + b_l\big) \odot (W_r \mathbf{Z} + b_r)\Big) + b_f
\end{equation}

and RMSNorm is a normalization layer \cite{zhang2019rootmeansquarelayer}.

Finally, a decoder maps back $\mathbf{Z}_L$ into a physical space $\mathbb{R}^3$ to predict the blood flow's velocity at the following time step (\textit{respectively} $\mathbb{R}^2$ to predict the fluid velocity in the case of the cylinder dataset). Both the encoder and the decoder are 2-layer Multi-layer perceptrons, where the second layer's width (\textit{respectively} the first layer for the decoder) is set to $d$. They also use an RMSNorm layer and a $\text{ReLU}$ activation function.

\subsection{Training}
\label{subsec:training}

We train our models using AdamW \cite{loshchilov2019decoupledweightdecayregularization}, with a warmup and cosine decay for the learning rate schedule (using an initial learning rate of $10^{-3}$). When switching a model to finer inputs, we compare two strategies:
\begin{enumerate}
    \item not resetting the learning rate or scheduling a new warmup, we simply resume the training with the same learning rate schedule.
    \item launching a new schedule for the remaining steps.
\end{enumerate}

\begin{figure}[ht!]
  \makebox[\textwidth][c]{\includegraphics[width=1\textwidth]{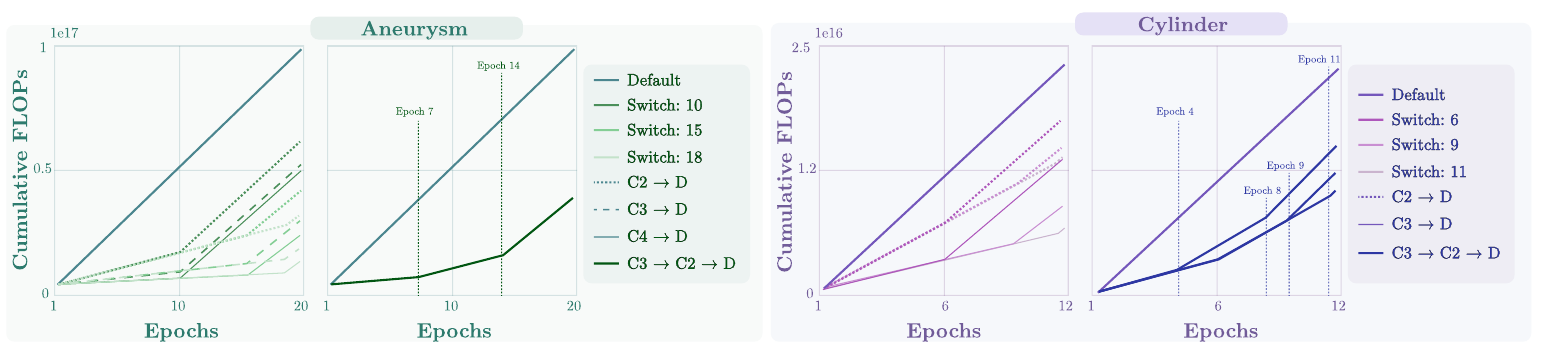}}%
  \caption{We present the curriculum used for both datasets in terms of cumulative FLOPs per training. For the more complex curriculum, we also detail when we switch between different datasets.}
  \label{fig:curriculm}
\end{figure}

During training and inference, inputs and outputs are normalized to zero-mean and unit variance. We use a Mean Squared Error (MSE) averaged on each node as a training loss for our model and add noise to our inputs to make it robust to error propagation over multiple time steps. We compare our models using a full auto-regressive prediction over the entire trajectories of the test set. 

Our models are trained for 12 epochs on the Cylinder dataset on a single L4 GPU and 20 epochs on the Aneurysm dataset on a single A100 GPU. We use similar curriculum for both datasets, as described in \autoref{fig:curriculm}, training for 50, 75 and 90\% of the time on a coarser dataset. We also experiment with more complex curriculum, using multiple coarse versions of datasets in a single training. In the remainder of the paper, we refer to the initial (or fine) dataset by \textbf{D}, and to its coarse version by \textbf{C}$\mathbf{n}$ where $\mathbf{n}$ refers to how coarse the dataset is. For example, $\textbf{C3} \rightarrow \textbf{D}$ means we start our training on the third version of the dataset before continuing on the fine version of the dataset.

It is important to note that while our model achieves a strong All-Rollout RMSE on the Cylinder dataset by default, it is not the case on the Aneurysm dataset. Even on twice as many epochs (20 to 40), our model reaches a plateau that it cannot break without curriculum learning (or a larger number of parameters).

\section{Results}
\label{sec:results}

\begin{figure}[ht!]
  \makebox[\textwidth][c]{\includegraphics[width=1\textwidth]{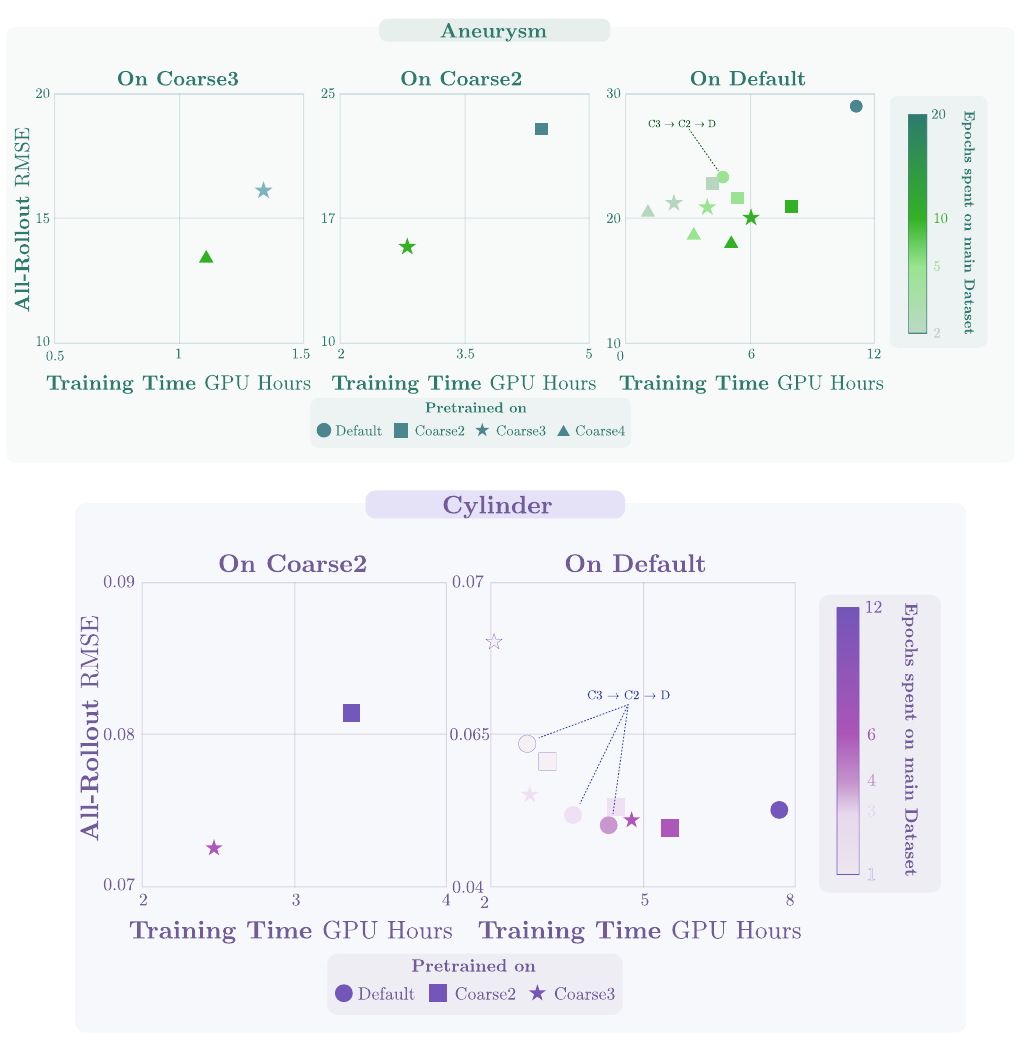}}%
  \caption{We present the All-Rollout RMSE of all trainings for both datasets, compared in terms of training time. We also display the version of the datasets on which the curriculum was created, as well as the duration for which each run was trained on the main dataset. We display the results on each version of the datasets that could be compared, with the main results on the plots labeled \emph{On Default}.}
  \label{fig:main-results}
\end{figure}

Across both datasets, starting on coarse meshes and progressively increasing resolution markedly reduces the training cost for a fixed number of epochs, while maintaining and often improving the final accuracy on the fine mesh. More importantly, on the Aneurysm dataset, it allows the model to properly learn the underlying physics and break through a plateau (see \autoref{fig:main-results}).

We find that using curriculum learning, even in its simplest form (a single phase pretraining on a coarser dataset), can cut the training time by up to 50\%. We also run experiments on the coarse datasets (for example, $\textbf{C3} \rightarrow \textbf{C2}$ and \textbf{C4} $\rightarrow \textbf{C3}$ on the Aneurysm dataset and $\textbf{C3} \rightarrow \textbf{C2}$ on the Cylinder dataset) and find that those runs also improve the All-rollout RMSE while reducing the training time.

On the Aneurysm dataset, we obtain the best results by pretraining on the coarsest dataset for 50\% of the training time. On the Cylinder dataset, best results are obtained on the least coarse version, also with 50\% of the training time. We attribute this difference to the overall mesh size discrepancies between the two datasets.

\begin{wrapfigure}{r}{0.545\textwidth}
\vspace{-40pt}
  \begin{center}
      \begin{tabular}{c}
        \includegraphics[width=1\linewidth]{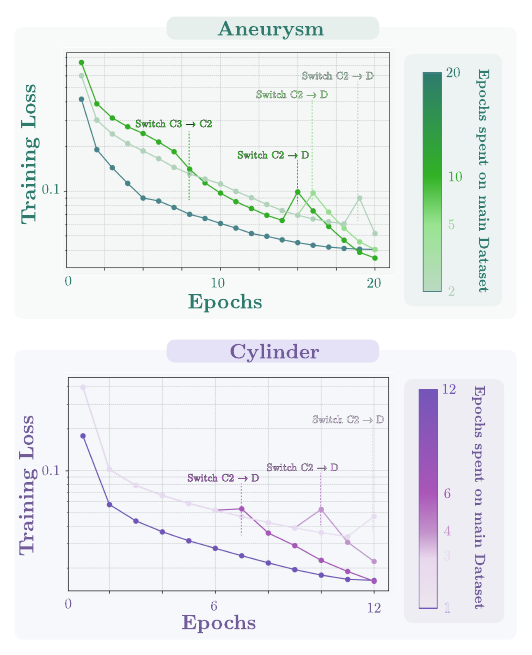}
    \end{tabular}
    \end{center}
    \caption{We display the training loss for several trainings, including the main one for both datasets. We also display when the dataset switching took place.}
    \label{fig:loss}
\vspace{-30pt}
\end{wrapfigure}

We also display the training loss for several training runs in \autoref{fig:loss}. First, we can see that the coarser the mesh, the higher the training loss. This is expected, given that the task becomes increasingly complex as we remove granularity. Second, we see once again that the earlier the switch, the better the performance. For aneurysms, curriculum trainings not only descend faster early on, they also reach lower loss for the \emph{same number of epochs spent on the finest mesh}, indicating that the coarse pretraining phase initializes the model in a region of parameter space from which the fine–mesh objective is easier to optimize. The cylinder curves show the same acceleration in the first few epochs after the switch.

\subsection{Ablation study}

\paragraph{Learning rate reset}

\begin{figure}[ht!]
  \makebox[\textwidth][c]{\includegraphics[width=1\textwidth]{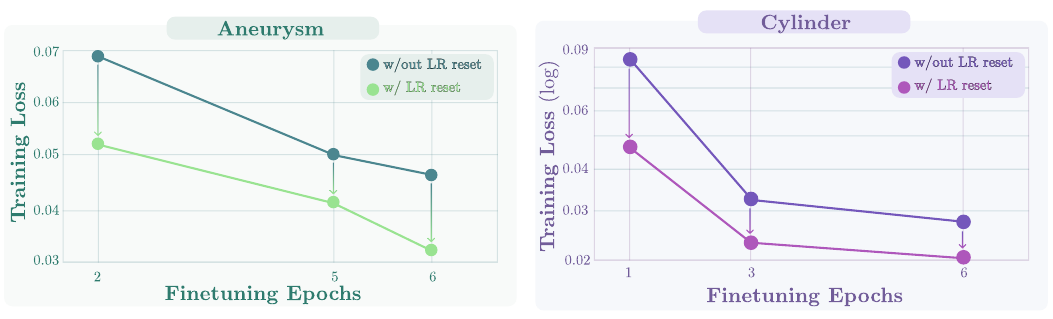}}%
  \caption{We display the final training loss for several curriculum on both datasets. Overall, we find that regardless of the approach, resetting the learning rate schedule (i.e., introducing a new warmup phase and a new cosine decay schedule) consistently improves performance.}
\label{fig:lr}
\end{figure}

We ablate the learning rate schedule when moving to a finer resolution in \autoref{fig:lr}. Resetting the learning rate upon each switch improves the first few epochs of fine–tuning on both datasets (sharper loss drops right after the switch), with a more pronounced effect on the aneurysm task. In our main experiments, we constantly reset the learning rate when switching between datasets. 

\paragraph{Curriculums}

\begin{figure}[ht!]
  \makebox[\textwidth][c]{\includegraphics[width=1\textwidth]{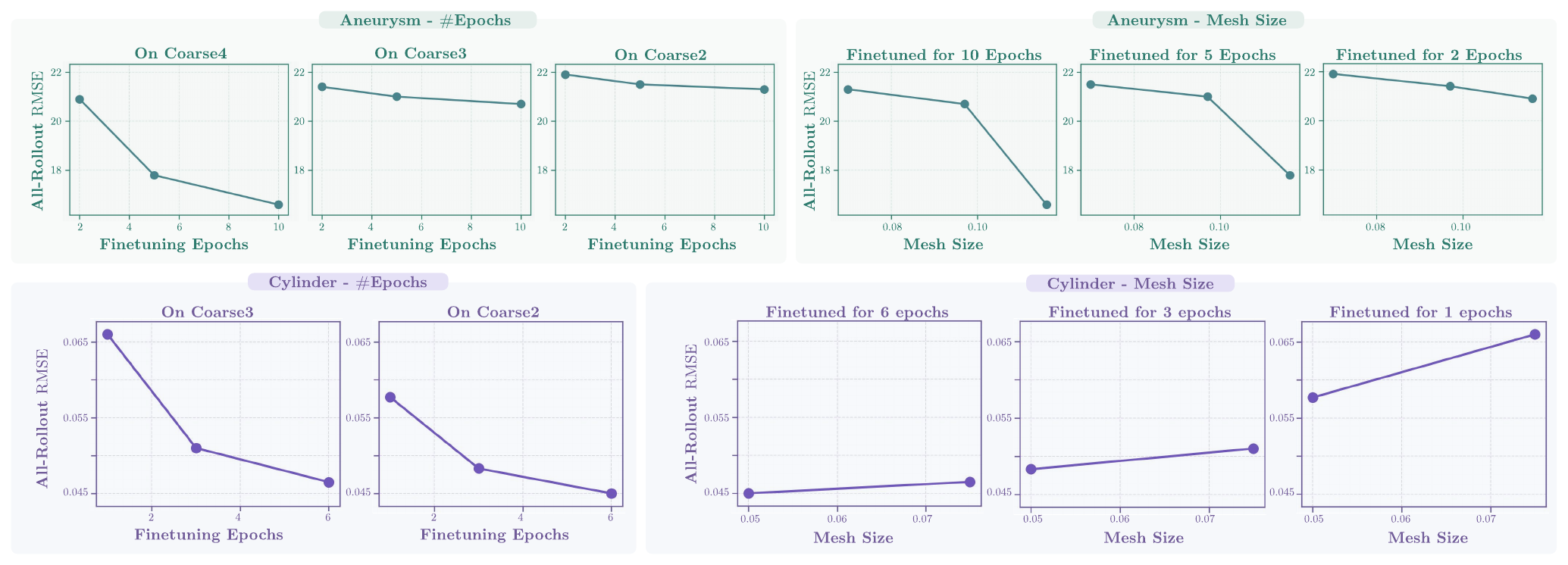}}%
  \caption{We display the All-rollout RMSE on both datasets based on either the mesh size of the pretraining phase, or the length of the finetuning phase.}
  \label{fig:ablation}
\end{figure}

Holding the pretraining mesh fixed, increasing fine–tuning epochs on the finest dataset monotonically improves all–rollout RMSE on both aneurysm and cylinder (see \autoref{fig:ablation}: top–left and bottom–left panels). We also find that the smaller the difference between the coarse dataset and the default one, the less impact the number of finetuning steps has. 

\paragraph{Mesh size}

On the Cylinder dataset, we find that the mesh size does not make a significant difference (see \autoref{fig:ablation}: top–right and bottom–right panels). This suggests that for a default dataset that does not hold a large number of vertices, simply reducing the number of nodes by a factor of 2 is enough to speed up the training time without a loss of performance.

On the Aneurysm dataset, however, we find that the larger the mesh size (\emph{i.e.,} the coarser the mesh), the better the performance, with increasing gains as the model is finetuned more. This suggests that on very refined datasets, aiming for a very coarse version of the meshes (even if they make less sense from a pure numerical simulation point of view) will allow for a significant speed-up in training with the same or even better performances.

\paragraph{Summary}

Overall, we find that no matter the mesh-based dataset used to train a GNN, it is always useful to split the training between the default dataset and a coarse version of it. Depending on the size of this default dataset, one can either: 
\begin{itemize}
    \item on small meshes (<10 thousand nodes): speed-up the training by up to 50\% for the same performances
    \item on medium meshes (<50 thousand nodes): speed up the training by up to 50\% for better performances
    \item on large meshes (>50 thousand nodes and up to 300 thousand nodes\footnote{while not mentioned in this paper, one can scale this approach to train a model even on very large meshes, up to 1 million of nodes}): successfully train a model that would otherwise fail, either because the dataset is too difficult or because the training time would otherwise be too impractical. 
\end{itemize}

Finally, a recipe could be summarized as follows: 

\begin{itemize}
    \item reset the learning rate schedule when switching between datasets
    \item use a simple approach, such as pretraining on a coarse version before finetuning on the main dataset
    \item the larger the main dataset, the coarser the dataset to pretrain on
    \item aim for 50\% of the total optimization steps on the default dataset. If impractical, aim for the most possible given the compute budget.
\end{itemize}

\section{Conclusion}

We have adapted and studied curriculum learning for GNN applied to mesh-based simulations. We find that, regardless of the approach, pretraining on coarse meshes yields similar or better results, while reducing the total training time by up to 50\%. This method is simple to implement, and even extremely coarse meshes that are unusable in terms of traditional numerical solvers can be utilized for such pretraining tasks, with no apparent drawbacks. 

\section{Acknowledgements}

\noindent The authors acknowledge the financial support from ERC grant no 2021-CoG-101045042, CURE. Views and opinions expressed are however those of the author(s) only and do not necessarily reflect those of the European Union or the European Research Council. Neither the European Union nor the granting authority can be held responsible for them.

\bibliography{template} 
\bibliographystyle{plain} 

\end{document}